\ifdefined\XeTeXrevision\else\pdfoutput=1\fi
\documentclass[11pt]{article}
\usepackage[margin=1in]{geometry}
\usepackage{amsmath,amssymb,amsthm,bm}
\usepackage{graphicx}
\usepackage{booktabs}
\usepackage[hidelinks]{hyperref}
\usepackage{microtype}
\newcommand{\citep}[1]{\cite{#1}}

\newtheorem{proposition}{Proposition}

\newtheorem{remark}{Remark}
\newcommand{\R}{\mathbb{R}}
\newcommand{\norm}[1]{\left\lVert #1 \right\rVert}
\newcommand{\E}{\mathbb{E}}
\newcommand{\Sq}{\Sigma_q}
\newcommand{\Sk}{\Sigma_k}

\title{SAKI-RoPE: Score-Aware Low-Rank Key Indexing\\with RoPE-Aware Covariance and Random-Matrix Noise Analysis}

\author{Lin Zhang\\
Data Science and Platform, Cotality\\
\texttt{linzhang1126@gmail.com}}
\date{October 2026}

\begin{document}
\maketitle

\begin{abstract}
Existing low-rank KV-cache indexes preserve either model weights or key
variance --- neither of which is the quantity attention actually uses. We
derive the expected attention-score distortion induced by key-side rank-$r$
compression and show it defines a two-sided covariance-weighted low-rank
approximation, $\mathcal{L}(P)=\|\Sq^{1/2}A(I-P)\Sk^{1/2}\|_F^2$, combining
the model's scoring operator $A = W_Q^\top W_K$ with the realized query
\emph{and} key statistics; a margin condition converts $\mathcal{L}$ control
into top-$k$ recall. Its rank-$r$ optimum has a closed-form asymmetric
factorization obtained from the SVD of the whitened query--key operator ---
an \emph{optimal rank-$r$ linear score map} that no span-restricted
(projector-based) method, PCA included, can express. The resulting
training-free index, \textbf{SAKI}, beats key-PCA at every rank on every
model tested --- LLaMA-3.1-8B, Qwen2.5-7B, Mistral-7B-v0.1, and
Llama-3.2-3B --- removing 13--30\% of PCA's remaining top-64 recall error
at $r{=}32$ (e.g.\ $0.748 \to 0.799$ on LLaMA-3.1-8B, $0.786 \to 0.850$ on
Qwen2.5-7B), improving 68--89\% of heads per model, with gains concentrated
in deep layers where query and key geometry diverge --- equivalently, PCA
needs 1.3--1.7$\times$ the rank to match SAKI's recall. A ranking-loss
refinement of the closed-form factors (SAKI-Rank) raises LLaMA-3.1-8B
recall to $0.835$, removing a third of PCA's remaining error and matching
PCA at half the rank. Because RoPE rotates keys by their distance from the query, we make the closed
form RoPE-aware by averaging the query covariance over rotations; this
\textbf{SAKI-RoPE}, our main method, raises LLaMA-3.1-8B recall to $0.861$, beats
PCA on every head of all four models, and needs only 28--39\% of PCA's rank to match it at $r{=}32$ (37--56\% at $r{=}64$) across the four models. These gains are consistent rather than dramatic, and they are
recall gains: we make no perplexity claim. In select-then-fetch perplexity the
differences between indexes are small and depend on model and text, and we
identify a failure mode of any low-rank index at long context: calibrated on
more than a few hundred tokens, it silently drops the attention-sink token,
which collapses 32K perplexity unless the sink is pinned. The contribution
is a principled objective with an exact solution, whose practical payoff is a
smaller index at equal recall. The same gap holds
against objective-level reimplementations of the 2026 data-aware family
(CARE-style activation-preserving factorization, STAR-KV-style adaptive
rank) and survives calibration-domain shift: SAKI wins all nine
calibrate$\to$evaluate pairs across literature, non-fiction, and code.
Because the index is
built from covariances estimated on finite, autocorrelated calibration
data, we treat their estimation noise with random-matrix theory: comparing
each head's spectrum against the Marchenko--Pastur noise bulk (with an
autocorrelation-corrected effective sample size) separates reliable signal
directions from sampling artifacts. This noise analysis yields three
practical facts --- SAKI needs only 512 calibration tokens to match PCA's
full-calibration recall; clipping the noise-bulk eigenvalues of both
covariances is a genuine correction when calibration is short (256 clipped
tokens match 512 raw ones), neutral when it is ample, and harmful to PCA
--- the correction is specific to the score-aware estimator; and
SAKI's per-head gains concentrate on heads whose key spectra sit mostly
\emph{inside} the noise bulk, exactly where variance offers PCA no
reliable signal. The theory predicts the
experiment: per-head predicted score-MSE
reduction matches measurement at Pearson $r = 0.997$ (median gap $0.0009$),
and our variant ordering --- exact asymmetric optimum $>$ weighted span $>$
raw span $\approx$ PCA --- shows the gain comes from the objective, not from
inserting covariances somewhere. An operator-geometry diagnosis (extreme,
cross-model non-normality of $A$; heavy-tailed weight spectra; hub-shaped
cross-subspace coupling) explains why weight-only and invariant-subspace
indexes fail, and why data-aware reconstruction was the right direction with
the wrong objective.
\end{abstract}

\section{Introduction}

At a million tokens of context, an 8B-parameter model's KV cache occupies
$\sim$130\,GB in bf16 --- decoding becomes a retrieval problem. Practical
systems separate \emph{selection} from \emph{reading}: a cheap index over
cached keys nominates candidates, exact values are fetched for the winners
\citep{quest,shadowkv}, and sparse architectures build the split into
training \citep{nsa}. The index is a rank-$r$ sketch of the keys; its
quality gates everything downstream.

What subspace should the index use? The field's answers fall into two
objective classes. \emph{Weight-side}: truncate the projection matrices ---
optimal for the operator, blind to data. \emph{Data-aware reconstruction}:
preserve the cached tensors under the activation distribution --- from key
PCA through its increasingly sophisticated 2026 descendants
\citep{kvcore,care,ojakv,starkv}. On a common protocol (all 1{,}024 heads of
LLaMA-3.1-8B, 4{,}096-token natural text, exact queries, top-64 recall) the
class representatives achieve $0.366$ (weight-SVD) and $0.748$ (key-PCA) at
$r{=}32$.

Neither class optimizes what attention uses. Scores are $s = q^\top k$;
retrieval needs their \emph{ranking}. PCA preserves
$\E\norm{k - PP^\top k}^2$ --- key variance: a high-variance direction that
queries never read wastes an index dimension, while a modest-variance
direction that $W_Q$ amplifies can dominate scores and be discarded. This
paper derives the objective that key compression should optimize, solves it
exactly, and measures the consequence.

\paragraph{Contributions.}
\begin{enumerate}
  \item \textbf{Problem formulation} (\S\ref{sec:objective}). The expected
  score distortion of key-side rank-$r$ compression is a two-sided
  covariance-weighted low-rank approximation
  $\mathcal{L}(P)=\|\Sq^{1/2}A(I-P)\Sk^{1/2}\|_F^2$, derived from the
  attention-score loss itself with both query and key statistics; a margin
  condition (Prop.~\ref{prop:margin}) links $\mathcal{L}$ to top-$k$
  recall. We claim no new linear algebra: the rank-constrained optimum
  follows from the classical full-rank-weights reduction
  \citep{eckart,izenman,srebro}. The claim is the \emph{identification} of
  this objective for attention retrieval, its exact specialization, and its
  empirical consequence.
  \item \textbf{Algorithm} (\S\ref{sec:saki}). The optimum is an
  \emph{optimal rank-$r$ linear score map} $M_r = \Sq^{-1/2} C_r \Sk^{-1/2}$
  --- an asymmetric low-rank bilinear factorization, generally \emph{not} a
  projector --- computable from two $d\times d$ calibration covariances per
  head, foldable into $W_Q, W_K$, training-free, with an RMT eigenvalue-
  clipping correction for short calibration. A ranking-loss refinement of
  the same factors (SAKI-Rank) is a one-shot calibration-time fit.
  \item \textbf{SAKI-RoPE, our main method} (\S\ref{sec:rope}). The objective is
  derived on un-rotated scores, whereas deployed scores are rotated by the
  query--key distance. Replacing $\Sq$ by its average over rotations keeps the
  closed form (the un-averaged version is SAKI, the ablation) and raises median recall@64 from $0.799$ to $0.861$ on
  LLaMA-3.1-8B ($0.850 \to 0.887$, $0.841 \to 0.893$, $0.797 \to 0.869$ on
  Qwen2.5-7B, Mistral-7B, Llama-3.2-3B); it beats PCA on every head of all
  four models and needs 28--39\% of PCA's rank to match PCA at $r{=}32$.
  \item \textbf{Evidence that theory predicts practice}
  (\S\ref{sec:experiments}). SAKI beats PCA at all three ranks on all four
  models tested (13--30\% of remaining recall error removed at $r{=}32$;
  68--89\% of heads improved per model), with the largest gains in deep
  layers. Predicted per-head score-MSE reduction matches measurement at
  Pearson $r{=}0.997$, directly validating the independence approximation
  the derivation uses. The variant ordering --- exact asymmetric optimum
  $>$ weighted span $>$ raw span $\approx$ PCA --- isolates the objective,
  not covariance insertion, as the source of gain. SAKI also wins all nine
  calibration-domain transfer cells and clears objective-level
  reimplementations of the 2026 data-aware family by $+0.044$ under a
  common protocol. SAKI-Rank reaches $0.835$ recall on LLaMA-3.1-8B,
  matching PCA at half the rank.
  \item \textbf{An attention-sink failure mode} (\S\ref{sec:experiments}).
  Low-rank indexes calibrated on long prefixes lose the attention-sink token
  (kept in 34--47\% of cases, against 87--99\% with a 512-token calibration),
  which collapses long-context perplexity (SAKI at 32K: $14.2 \to 64.3$ when
  the calibration grows from 512 to 4{,}096 tokens); pinning the first four
  tokens removes the failure, and the RoPE-aware covariance avoids it without
  pinning.
  \item \textbf{Diagnosis} (\S\ref{sec:diagnosis}, details in appendix).
  A cross-model operator-geometry analysis --- extreme non-normality of
  $A_h$ (median Henrici $0.95$--$0.98$ across five models), heavy-tailed
  weight spectra, hub-shaped cross-subspace coupling --- explains why
  weight-only and invariant-subspace indexes fail and motivates the
  objective-level fix.
\end{enumerate}

\paragraph{What we claim, and what we do not.} The empirical gains are
\emph{consistent and moderate}, and we state them as such. SAKI improves
median recall@64 by roughly 3--7 points (13--30\% of PCA's remaining error at
$r{=}32$, depending on the model), and it does so in every one of the 12
(model, rank) cells, at every context length from 4K to 64K, and in all nine
calibration-domain pairs. The most useful reading is \emph{index memory}:
SAKI reaches the recall PCA attains at 1.3--1.7$\times$ the rank (equal
recall at 59--77\% of the index dimensions), and SAKI-Rank at about half. The RoPE-aware variant (\S\ref{sec:rope}) lifts the gap to PCA to 7--11 recall points on the same four models.
We do \emph{not} claim any gain in generation quality: in
select-then-fetch perplexity the selection budget $k'$ matters far more than
the choice of index; with the attention sink pinned, SAKI and PCA differ by a
few tenths of a perplexity point at most in the favourable cases, in a
direction that depends on the model and the text (Appendix~\ref{app:ppl}), and
no task-level benchmark is reported. The primary claim is therefore
the \emph{problem formulation and its exact solution} --- a score-aware
objective whose closed-form optimum predicts the measured per-head score
error to $r{=}0.997$ --- and its practical payoff is a consistently smaller
index at equal recall.

\section{Related Work}
\paragraph{Token selection and sparse attention.} H2O \citep{h2o}, SnapKV
\citep{snapkv} (eviction); Quest \citep{quest} (query-aware page bounds);
NSA \citep{nsa} and DeepSeek-V3.2's lightning indexer (learned selectors).
\paragraph{Channel-axis compression.} MLA \citep{mla} trains a latent cache;
Palu \citep{palu}, Eigen Attention \citep{eigenattention}, LoRC \citep{lorc}
apply SVD-family projections; ShadowKV \citep{shadowkv} sketches pre-RoPE
keys with data SVD and fetches exact values. Quantization \citep{kivi,
kvquant} composes with SAKI codes.
\paragraph{Contemporary data-aware KV compression.} A rapidly moving
2025--2026 line makes activation-awareness standard. KV-CoRE \citep{kvcore}
benchmarks data-dependent compressibility via the Frobenius-optimal
low-rank approximation of the cache (effective rank as the metric);
CARE \citep{care} converts GQA to MLA with an \emph{activation-preserving}
factorization that aligns the weight approximation with input activations
rather than weights alone, plus cross-layer rank allocation; OjaKV
\citep{ojakv} keeps the first and most recent tokens full-rank and adapts
the projection basis online with Oja's rule to survive distribution shift;
STAR-KV \citep{starkv} learns per-head/per-block ranks by differentiable
soft thresholding, with a hybrid K/V decomposition, quantization, and
Triton kernels. ``Data-aware beats weight-SVD'' is therefore \emph{not} a
claim of this paper: every method in this family --- however adaptive its
rank or basis --- optimizes \emph{reconstruction} of the cached tensors
under the activation distribution. In the index setting their objectives
reduce to activation-weighted key reconstruction, which we run
head-to-head under our protocol as objective-level reimplementations
(\S\ref{sec:experiments}: CARE-style activation-preserving factorization,
STAR-KV-style matched-budget adaptive rank, with full-data PCA as the
in-domain upper bound of OjaKV's online estimate). The distinction from
CARE, the closest neighbor, is explicit: \emph{CARE optimizes
representation fidelity under attention-module conversion; SAKI optimizes
pairwise attention-score fidelity for a retrieval index, with asymmetric
query/key factors derived from both covariances.} The axes compose:
OjaKV-style online tracking applies verbatim to $\Sq, \Sk$; STAR-KV-style
adaptive ranks to $r$ (and we test the latter).
\paragraph{Classical multivariate analysis and weighted low-rank
approximation.} The machinery is classical and we position against it
explicitly. Rank-constrained approximation under full-rank two-sided
weights reduces by substitution to Eckart--Young \citep{eckart}; this is
the solution structure of reduced-rank regression \citep{izenman}, with the
general weighted problem studied in \citep{srebro}. Canonical correlation
analysis \citep{hotelling} whitens the \emph{cross-covariance} of two
observed variable sets ($\Sigma_q^{-1/2}\Sigma_{qk}\Sigma_k^{-1/2}$); our
object differs in kind --- a \emph{fixed model operator} weighted (not
whitened) by marginal square roots, a known bilinear form rather than an
estimated cross-covariance. In retrieval, asymmetric MIPS \citep{alsh} and
ScaNN's anisotropic quantization \citep{scann} adopt inner-product-aware
objectives for \emph{corpus quantization} against generic queries; SAKI
derives a per-head \emph{subspace} through a fixed bilinear operator with
both distributions estimated from the model's own activations, foldable
into its weights. The novelty claim is deliberately narrow: \emph{new
problem formulation + exact specialization + empirical consequence}, not
new linear algebra.
\paragraph{Non-normal operator theory.} Transient growth and non-normal
dynamics are studied in fluids, RNNs, and optimization; we are unaware of
prior measurement of transformer score-operator non-normality
(\S\ref{sec:diagnosis}).

\section{Setup and baselines}
\label{sec:setup}
Head $h$ scores position $j$ against query $i$ via $s_{ij} = q_i^\top k_j =
x_i^\top A_h x_j$, $A_h = W_Q^{(h)\top} W_K^{(g)} \in \R^{D\times D}$, rank
$\le d$ (GQA; $D{=}4096$, $d{=}128$). The index compresses cached keys to
$r \ll d$ dims; queries are exact at decode; codes are pre-RoPE with
rotation applied after reconstruction \citep{shadowkv}. Protocol: exact
float32 streaming forward pass of LLaMA-3.1-8B on 4{,}096 tokens of natural
text; recall of the true top-64 attended positions over the last 512
queries; medians across all 1{,}024 heads.

\begin{table}[h]
\centering
\begin{tabular}{lccc}
\toprule
Index & $r=16$ & $r=32$ & $r=64$ \\
\midrule
\textbf{SAKI-opt (exact $M_r$, ours)} & \textbf{0.731} & \textbf{0.799} & \textbf{0.876} \\
SAP-map (weighted span, ours) & 0.696 & 0.766 & 0.845 \\
SAP-svd (singular span, ours) & 0.669 & 0.740 & 0.829 \\
key PCA & 0.679 & 0.748 & 0.833 \\
weight SVD & 0.267 & 0.366 & 0.555 \\
Schur invariant subspace & 0.235 & 0.319 & 0.411 \\
\bottomrule
\end{tabular}
\caption{Median top-64 attention recall, 1{,}024 heads, context 4{,}096.
Relative to PCA's remaining error, SAKI-opt removes $16\%$/$20\%$/$26\%$
at $r = 16/32/64$. Bootstrap 95\% intervals on the median over heads
($B{=}2000$) are $\pm 0.005$ or tighter for every cell; the SAKI$-$PCA
median delta is $+0.050\,[+0.046, +0.056]$, $+0.041\,[+0.036, +0.045]$,
$+0.033\,[+0.030, +0.035]$ at $r=16/32/64$.}
\label{tab:main}
\end{table}

\section{Diagnosis: why weight geometry is not enough}
\label{sec:diagnosis}
Three measured facts, established across five open models and detailed in
Appendix~\ref{app:atlas}, motivate the objective-level fix. \textbf{(a)}
Attention score operators are \emph{extremely non-normal}: median Henrici
departure $0.975$--$0.984$ for every standard-attention model measured
(LLaMA-3.1/3.2, Qwen2.5, Mistral), $\sim$88\% complex eigenvalues, flat in
depth; the MLA-based DeepSeek-V2-Lite is measurably closer to normal
($0.948$) --- latent bottlenecks regularize score-operator geometry.
Singular and invariant subspaces are therefore nearly disjoint objects, and
``which subspace'' is a modeling decision. \textbf{(b)} Weight spectra are
\emph{heavy-tailed} (median stable rank $27.7/128$ against a 99\%-energy
rank of $106/128$): weight-only truncation at useful ranks discards live
directions --- hence weight-SVD's $0.366$. \textbf{(c)} Half the operator's
energy lives in \emph{hub-shaped cross-subspace couplings} (ordered Schur
form): key-side truncation to a leading invariant subspace severs exactly
those couplings, predicting --- correctly, see Table~\ref{tab:main} ---
that eigen/Jordan-style compression fails hardest. PCA's strength shows
activation geometry dominates weight geometry; its residual weakness is the
subject of the next section.

\section{The score-aware objective and its exact solution}
\label{sec:objective}
Let key-side compression be a rank-$r$ linear map $P$ applied to cached
keys. The score error is $e = x_q^\top A (I-P) x_k$. Under an independent
query/key model with second moments $\Sq^{x} = \E[x_q x_q^\top]$,
$\Sk^{x} = \E[x_k x_k^\top]$,
\begin{equation}
\E[e^2]
= \operatorname{tr}\!\left[ A (I-P)\, \Sk^{x} (I-P)^\top A^\top\, \Sq^{x} \right]
= \norm{ (\Sq^{x})^{1/2} A (I - P)\, (\Sk^{x})^{1/2} }_F^2
\;=:\; \mathcal{L}(P).
\label{eq:objective}
\end{equation}
PCA minimizes $\norm{(I-P)(\Sk^{x})^{1/2}}_F^2$; the correct weight is
$\Sq^{1/2} A (\cdot)\, \Sk^{1/2}$. In head coordinates ($q = W_Q x$,
$k = W_K x$) the operator is absorbed into the query moment: with
$\Sq = \E[qq^\top] = W_Q \Sigma_x W_Q^\top$ and $\Sk = \E[kk^\top]$,
\begin{equation}
\mathcal{L}(P) = \norm{ \Sq^{1/2} (I - P)\, \Sk^{1/2} }_F^2 ,
\qquad
C := \Sq^{1/2} \Sk^{1/2} \in \R^{d \times d},
\label{eq:headspace}
\end{equation}
computable from two $d\times d$ calibration covariances per head. $\Sq$ is
where the attention operator lives: key variance $\times$ query
probability $\times$ scoring geometry --- the quantity retrieval needs.

\begin{proposition}[Exact minimizer over rank-$r$ maps; classical reduction]
\label{prop:sap}
Over all rank-$r$ linear maps $P: \R^d \to \R^d$,
$\mathcal{L}$ in Eq.~(\ref{eq:headspace}) is minimized by
\begin{equation}
M_r \;=\; \Sq^{-1/2}\, C_r\, \Sk^{-1/2},
\qquad C_r = U_r \Lambda_r V_r^\top \ \text{the truncated SVD of } C,
\label{eq:mr}
\end{equation}
with optimum $\mathcal{L}(M_r) = \sum_{i>r} \lambda_i(C)^2$ (substitution
$\tilde P = \Sq^{1/2} P \Sk^{1/2}$ plus Eckart--Young; inverses read as
regularized pseudoinverses when covariances are singular ---
Appendix~\ref{app:proof} gives the full derivation including centering,
rank-deficiency, and finite-sample regularization).
\end{proposition}

\begin{remark}[Terminology]
$M_r$ is generally \emph{not} idempotent ($M_r^2 \ne M_r$): it is an
\emph{optimal rank-$r$ linear score map} --- an asymmetric low-rank bilinear
factorization --- not an oblique projector. The distinction matters for the
result: orthogonal projectors (all span-based methods, PCA included) form a
strict subset of rank-$r$ maps, so the optimum over the larger class can
only be better, and Table~\ref{tab:main} shows the strict-subset gap is
real ($0.799$ vs.\ $0.766$/$0.740$ at $r{=}32$).
\end{remark}

\begin{proposition}[Top-$k$ preservation]
\label{prop:margin}
If $|s_{ij} - \hat s_{ij}| \le \varepsilon$ for all $j$ and the true
top-$k$ margin is $\gamma_k = s_{(k)} - s_{(k+1)}$, the top-$k$ set is
recovered exactly when $2\varepsilon < \gamma_k$; recall degrades
continuously in $\varepsilon / \gamma_k$. Control of $\mathcal{L}$
(mean-square $\varepsilon$) therefore transfers to recall.
\end{proposition}

\begin{remark}[The independence assumption, tested]
\label{rem:indep}
In self-attention $q_i$ and cached $k_j$ come from the same sequence, with
positional and content correlations and RoPE's relative rotation;
Eq.~(\ref{eq:objective}) with a general cross-moment adds the term
$-2\operatorname{tr}[A(I-P)\,\Sigma_{kq}\,A(I-P)]$-type corrections that
have no closed form. Rather than assume the approximation harmless, we
measure it: per head, the reduction in score-MSE predicted by
Prop.~\ref{prop:sap} ($1 - \sum_{i>r}\lambda_i^2 / \sum_i \lambda_i^2$)
against the actual reduction on real causal query/key pairs. Across all
1{,}024 heads at $r{=}32$: Pearson $r = 0.997$, median
$|$predicted $-$ actual$| = 0.0009$ (Appendix~\ref{app:ablation}). For
score-MSE on natural text, the independence model is near-exact; RoPE
handling (pre-rotation codes) remains the approximation to refine.
\end{remark}

\section{SAKI}
\label{sec:saki}
Per head, from one calibration pass: $\Sq$ (uncentered second moment of
pre-RoPE queries), $\Sk$ and mean $\mu$ (centered keys); form $C$; store
\begin{equation}
\text{key code } c_j = B_k^\top (k_j - \mu) \in \R^r, \quad
B_k = \Sk^{-1/2} V_r \Lambda_r^{1/2}; \qquad
B_q = \Sq^{-1/2} U_r \Lambda_r^{1/2},
\end{equation}
so that $\hat s = (B_q^\top q)^\top c_j + q^\top \mu$ reproduces
$q^\top M_r (k - \mu) + q^\top \mu$ exactly
(Appendix~\ref{app:proof} verifies the factorization). Both maps fold into
$W_K, W_Q$ offline; when composing with RoPE, reconstruct
$\hat k = \mu + M_r(k - \mu)$ and rotate. Cost: two $d \times d$
covariances and one $d \times d$ SVD per head --- seconds per model.
Variants isolating what matters: \textbf{SAKI-opt} (exact $M_r$),
\textbf{SAP-map} ($\Sk^{1/2}$-weighted span, orthonormalized),
\textbf{SAP-svd} (raw right-singular span of $C$).

\subsection{RoPE-aware covariance}
\label{sec:rope}
The objective is derived on pre-rotation scores $q^\top k$, but the score a
head actually computes between a query at position $m$ and a key at position
$n \le m$ is $q^\top R_{-\Delta}\,k$ with $\Delta = m-n \ge 0$ and $R_\Delta$ the
block-diagonal RoPE rotation by $\Delta$ positions. The expected distortion of
a linear score map $M$ at distance $\Delta$ is
$\mathrm{tr}\big(R_\Delta \Sq R_\Delta^\top (I-M)\Sk(I-M)^\top\big)$, i.e.\ the
same objective with $\Sq$ replaced by the rotated covariance
$R_\Delta \Sq R_\Delta^\top$. Averaging over the distances present in a context
of $N$ tokens,
\begin{equation}
\bar\Sigma_q \;=\; \frac1N \sum_{\Delta=0}^{N-1} R_\Delta\, \Sq\, R_\Delta^\top ,
\end{equation}
all of the derivation applies verbatim with $\Sq \to \bar\Sigma_q$: the
optimum is $M_r = \bar\Sigma_q^{-1/2} \bar C_r \Sk^{-1/2}$ with
$\bar C = \bar\Sigma_q^{1/2}\Sk^{1/2}$. Fast RoPE planes rotate through many
cycles and average to an isotropic block, while the slowest planes barely
rotate and keep their structure, so the averaged covariance preserves what
matters at \emph{every} distance instead of fitting distance zero. We use 128
evenly spaced $\Delta$ per head ($128$ products of $d\times d$ matrices); the
rest of the pipeline is unchanged. We call the resulting index \textbf{SAKI-RoPE}; it is our main method, and the un-averaged closed form (SAKI) is its ablation and baseline.

\section{Experiments}
\label{sec:experiments}

\begin{figure}[t]
\centering
\includegraphics[width=\linewidth]{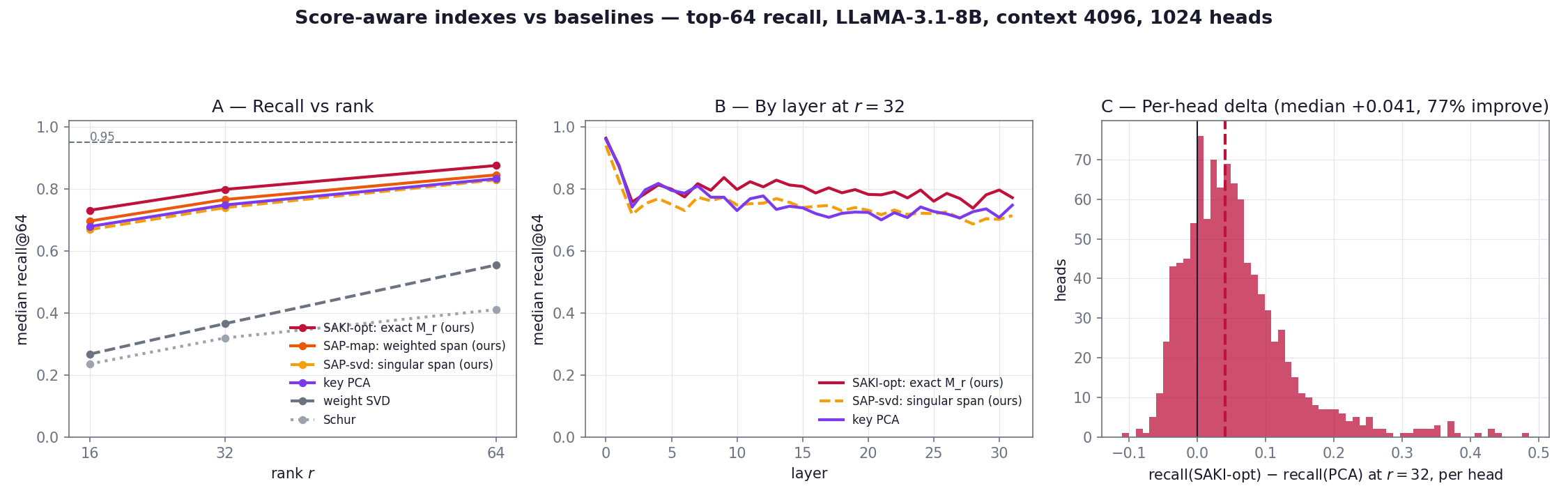}
\caption{Score-aware indexes vs.\ baselines. (A)~Recall vs.\ rank.
(B)~By layer at $r{=}32$: SAKI's gains concentrate in deep layers where
PCA is weakest. (C)~Per-head delta at $r{=}32$: median $+0.041$, $76.9\%$
of heads improve.}
\label{fig:sap}
\end{figure}

\begin{figure}[t]
\centering
\includegraphics[width=\linewidth]{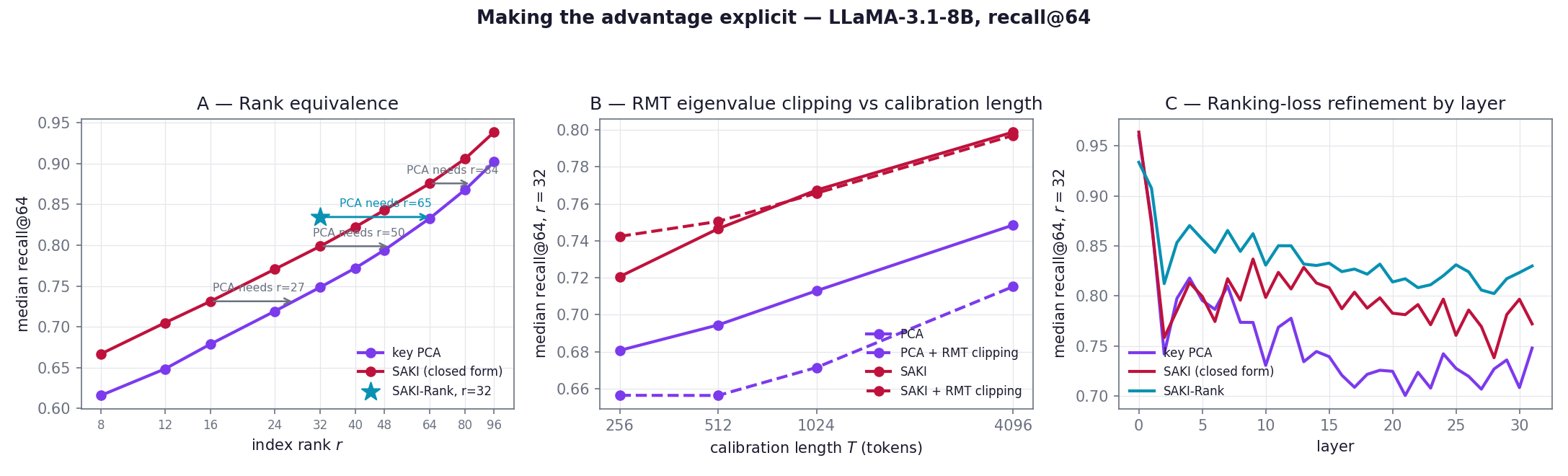}
\caption{(A)~Rank equivalence: SAKI reaches PCA's recall at 59--77\% of the
rank; SAKI-Rank at $r{=}32$ matches PCA at $r{\approx}65$. (B)~RMT eigenvalue
clipping vs.\ calibration length: a short-$T$ correction for SAKI, harmful
for PCA. (C)~SAKI-Rank by layer.}
\label{fig:adv}
\end{figure}

\paragraph{Main result.} SAKI-opt beats PCA at every rank
(Table~\ref{tab:main}, Fig.~\ref{fig:sap}); framed against what remains,
it removes $16\%$, $20\%$, and $26\%$ of PCA's residual recall error at
$r = 16, 32, 64$. The per-head delta at $r{=}32$ has median $+0.041$ with
$76.9\%$ of heads improving ($79.3\%$ at $r{=}64$), and the gains
concentrate in layers 16--31 ($+0.061$ median) --- precisely where query
and key geometry diverge and a query-blind subspace is costliest.
\paragraph{Rank equivalence.} The same gap stated in memory: on a fine rank
grid (Fig.~\ref{fig:adv}A), SAKI at $r = 16/32/64$ reaches the recall PCA
attains only at $r = 27/50/84$ --- equal recall at 59--77\% of the index
rank, i.e.\ of the resident key-code memory.
\paragraph{The objective, not the covariances, wins.} The variant ordering
is exact-optimum $>$ weighted span $>$ raw span $\approx$ PCA at every
rank. Merely inserting covariance matrices into a span construction
(SAP-svd) does not beat PCA; solving the derived objective over the full
class of rank-$r$ maps does. This is the empirical signature of the
asymmetric factorization.
\paragraph{Theory predicts practice.} The predicted-vs-actual score-MSE
ablation (Remark~\ref{rem:indep}, Appendix~\ref{app:ablation}) shows per-head
agreement at $r = 0.997$ across 1{,}024 heads --- the derivation's
independence model describes real attention statistics almost exactly,
so the closed form can be trusted as the deployment-time optimum for its
objective.
\paragraph{Cross-model replication.} Running the identical pipeline (exact
streaming forward pass on the same calibration text, PCA vs.\ exact $M_r$)
on three further models --- including Qwen2.5-7B, whose attention carries
QKV biases, and Llama-3.2-3B with factor-32 llama3 RoPE scaling ---
replicates the result at every rank on every model
(Table~\ref{tab:xmodelsaki}). The gain is largest on Qwen2.5-7B (30\% of
remaining error removed at $r{=}32$; deep-layer deltas up to $+0.12$; 89\%
of heads improve) and smallest but still uniform on Mistral-7B (13\%; 71\%
of heads). SAKI never loses to PCA in any (model, rank) cell.

\begin{table}[h]
\centering
\begin{tabular}{lccccc}
\toprule
Model & \multicolumn{3}{c}{PCA $\to$ SAKI (median recall@64)} & heads$\uparrow$ & err.\ removed \\
 & $r=16$ & $r=32$ & $r=64$ & @$r{=}32$ & @$r{=}32$ \\
\midrule
LLaMA-3.1-8B & .679$\to$.731 & .748$\to$.799 & .833$\to$.876 & 76.9\% & 20\% \\
Qwen2.5-7B & .717$\to$.777 & .786$\to$.850 & .864$\to$.914 & 89.0\% & 30\% \\
Mistral-7B-v0.1 & .742$\to$.774 & .818$\to$.841 & .900$\to$.914 & 71.1\% & 13\% \\
Llama-3.2-3B & .679$\to$.721 & .760$\to$.797 & .850$\to$.874 & 68.4\% & 15\% \\
\bottomrule
\end{tabular}
\caption{Cross-model SAKI: identical protocol on four models. SAKI beats
PCA in every cell.}
\label{tab:xmodelsaki}
\end{table}

\paragraph{RoPE-aware SAKI.} Replacing $\Sq$ by $\bar\Sigma_q$
(\S\ref{sec:rope}) lifts recall further on every model (Table~\ref{tab:xmodelrope}):
LLaMA-3.1-8B $0.799 \to 0.861$, Qwen2.5-7B $0.850 \to 0.887$, Mistral-7B
$0.841 \to 0.893$, Llama-3.2-3B $0.797 \to 0.869$, and it beats PCA on
\emph{every} head of all four models (against 68--89\% for the plain closed
form). It is never worse than plain SAKI in the 1{,}024 heads of LLaMA-3.1-8B
($5$ of $1{,}024$ on Mistral-7B). The gain is largest where the plain closed
form was weakest: in layers 0--7 of LLaMA-3.1-8B plain SAKI only ties PCA
($+0.002$) while SAKI-RoPE leads by $0.077$; in the three deeper groups of
eight layers the lead over PCA grows from $+0.061/+0.062/+0.040$ to
$+0.113/+0.121/+0.109$. At equal recall (Tables~\ref{tab:rankrope} and~\ref{tab:rankmodels}) SAKI-RoPE needs
$28\%$ of PCA's rank to match PCA at $r{=}32$ on LLaMA-3.1-8B ($r \approx 9$) and
$37\%$ to match PCA at $r{=}64$ ($r \approx 24$), against $60$--$69\%$ for plain SAKI;
across the four models the figures are $28$--$39\%$ and $37$--$56\%$ (plain SAKI:
$55$--$79\%$ and $59$--$89\%$). With 512 calibration tokens the picture is the same
($0.694 \to 0.746 \to 0.816$ for PCA, SAKI, SAKI-RoPE on LLaMA-3.1-8B). The
mechanism is visible in the score error resolved by key distance
(LLaMA-3.1-8B, 32K context, 72 heads, full-prefix calibration, error relative
to the mean squared exact score): plain SAKI's error grows from $0.0008$ below
distance 16 to $0.0050$ beyond 16K, overtaking PCA's roughly flat
$0.0025$--$0.0032$, whereas SAKI-RoPE stays at $0.0009$--$0.0013$ at every
distance. At 32K the ordering is unchanged ($0.739$, $0.780$, $0.857$ for PCA,
SAKI, SAKI-RoPE on 72 LLaMA heads; $0.758$, $0.828$, $0.870$ on 224 Qwen
heads). RMT clipping and rotation averaging address overlapping problems:
clipping a SAKI-RoPE covariance lowers recall ($0.831 \to 0.790$ at 4K with
512 tokens, 72 heads), so we do not stack them. These are recall results; they
do not produce a perplexity difference we can resolve (below).

\begin{table}[t]
\centering
\small
\begin{tabular}{lcccc}
\toprule
Model & PCA & SAKI & SAKI-RoPE & heads $>$ PCA \\
\midrule
LLaMA-3.1-8B & .748 & .799 & \textbf{.861} & 77\% $\to$ 100\% \\
Qwen2.5-7B & .786 & .850 & \textbf{.887} & 89\% $\to$ 100\% \\
Mistral-7B-v0.1 & .818 & .841 & \textbf{.893} & 71\% $\to$ 100\% \\
Llama-3.2-3B & .760 & .797 & \textbf{.869} & 68\% $\to$ 100\% \\
\bottomrule
\end{tabular}
\caption{Median recall@64 at $r{=}32$ (4K context, full-sequence calibration, all heads: 1{,}024 / 784 / 1{,}024 / 672) and share of heads that beat PCA (SAKI $\to$ SAKI-RoPE).}
\label{tab:xmodelrope}
\end{table}

\begin{table}[t]
\centering
\small
\begin{tabular}{lccc}
\toprule
$r$ & PCA & SAKI & SAKI-RoPE \\
\midrule
8 & .616 & .667 & \textbf{.742} \\
12 & .648 & .705 & \textbf{.774} \\
16 & .679 & .731 & \textbf{.800} \\
24 & .719 & .770 & \textbf{.835} \\
32 & .748 & .799 & \textbf{.861} \\
48 & .794 & .843 & \textbf{.899} \\
64 & .833 & .876 & \textbf{.927} \\
\bottomrule
\end{tabular}
\caption{Median recall@64 by index rank (LLaMA-3.1-8B, 4K context, all 1{,}024 heads, full-sequence calibration). SAKI-RoPE at $r{\approx}9$ matches PCA at $r{=}32$ and at $r{\approx}24$ matches PCA at $r{=}64$ (interpolated); plain SAKI needs $60$--$69\%$ of PCA's rank.}
\label{tab:rankrope}
\end{table}

\begin{table}[t]
\centering
\scriptsize
\begin{tabular}{lccclcl}
\toprule
 & \multicolumn{3}{c}{match PCA at $r{=}32$} & \multicolumn{3}{c}{match PCA at $r{=}64$} \\
model & PCA recall & SAKI & SAKI-RoPE & PCA recall & SAKI & SAKI-RoPE \\
\midrule
LLaMA-3.1-8B & .748 & 60\% ($r{\approx}19$) & \textbf{28\% ($r{\approx}9$)} & .833 & 69\% ($r{\approx}44$) & \textbf{37\% ($r{\approx}24$)} \\
Qwen2.5-7B & .786 & 55\% ($r{\approx}18$) & \textbf{31\% ($r{\approx}10$)} & .864 & 59\% ($r{\approx}38$) & \textbf{37\% ($r{\approx}24$)} \\
Mistral-7B-v0.1 & .818 & 79\% ($r{\approx}25$) & \textbf{39\% ($r{\approx}13$)} & .900 & 89\% ($r{\approx}57$) & \textbf{56\% ($r{\approx}36$)} \\
Llama-3.2-3B & .760 & 71\% ($r{\approx}23$) & \textbf{34\% ($r{\approx}11$)} & .850 & 81\% ($r{\approx}52$) & \textbf{42\% ($r{\approx}27$)} \\
\bottomrule
\end{tabular}
\caption{Index rank needed to match PCA's recall, as a percentage of PCA's rank (rank in parentheses; linear interpolation of the median recall@64 over a rank grid from 6 to 64; 4K context, all heads, full-sequence calibration).}
\label{tab:rankmodels}
\end{table}

\begin{table}[t]
\centering
\scriptsize
\begin{tabular}{lcccc}
\toprule
model & $r{=}8$ & $r{=}16$ & $r{=}32$ & $r{=}64$ \\
\midrule
LLaMA-3.1-8B & .616 / .667 / \textbf{.742} & .679 / .731 / \textbf{.800} & .748 / .799 / \textbf{.861} & .833 / .876 / \textbf{.927} \\
Qwen2.5-7B & .660 / .714 / \textbf{.769} & .717 / .777 / \textbf{.828} & .786 / .850 / \textbf{.887} & .864 / .914 / \textbf{.941} \\
Mistral-7B-v0.1 & .666 / .701 / \textbf{.784} & .742 / .774 / \textbf{.836} & .818 / .841 / \textbf{.893} & .900 / .914 / \textbf{.945} \\
Llama-3.2-3B & .614 / .641 / \textbf{.735} & .679 / .721 / \textbf{.800} & .760 / .797 / \textbf{.869} & .850 / .874 / \textbf{.932} \\
\bottomrule
\end{tabular}
\caption{Median recall@64 by index rank, PCA / SAKI / SAKI-RoPE (4K context, all heads of each model, full-sequence calibration).}
\label{tab:ranksweep}
\end{table}

\begin{figure}[t]
\centering
\includegraphics[width=\linewidth]{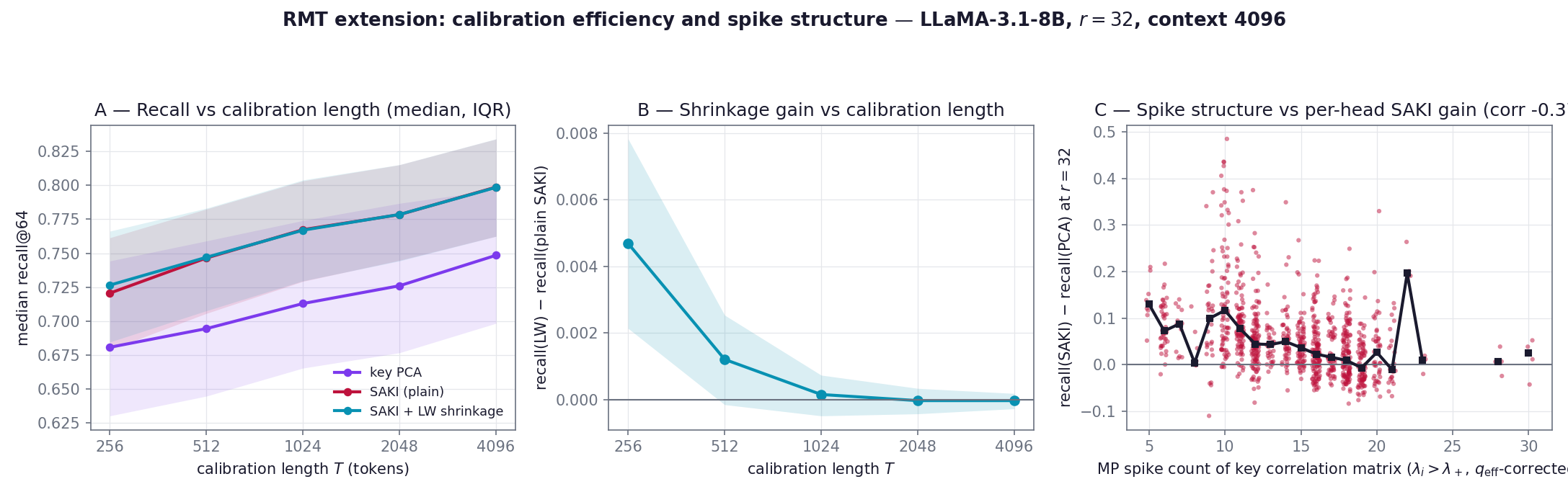}
\caption{RMT extension at $r{=}32$. (A)~Recall vs.\ calibration length:
SAKI at $T{=}512$ matches PCA at $T{=}4096$. (B)~Ledoit--Wolf shrinkage
helps only marginally at the smallest $T$. (C)~Per-head SAKI$-$PCA gain
vs.\ MP spike count of the key correlation matrix: gains concentrate on
bulk-dominated heads.}
\label{fig:rmt}
\end{figure}

\paragraph{Calibration efficiency, and a random-matrix view of estimation
noise.} PCA supplies eigenvectors; random-matrix theory judges which sample
eigenvalues are indistinguishable from estimation noise
\citep{mp,laloux}. Both $\Sq$ and $\Sk$ are sample second moments
($d{=}128$) whose estimation noise is governed by
$q_{\mathrm{eff}} = d / T_{\mathrm{eff}}$, with
$T_{\mathrm{eff}} = T/\tau_{\mathrm{int}}$ correcting for token
autocorrelation (measured median $\tau_{\mathrm{int}} = 3.2$). Sweeping the
calibration length $T$ (Fig.~\ref{fig:rmt}A): SAKI degrades gracefully ---
\textbf{calibrated on 512 tokens it matches PCA calibrated on 4{,}096}
($0.746$ vs.\ $0.748$), an $\sim$8$\times$ calibration-efficiency
advantage, and it beats same-$T$ PCA by $+0.04$--$0.05$ at every $T$.
Two corrections were tested. Ledoit--Wolf shrinkage \citep{ledoitwolf} of
both moments adds only $+0.005$ at $T{=}256$ and nil beyond
(Fig.~\ref{fig:rmt}B). \textbf{RMT eigenvalue clipping} \citep{laloux} ---
replacing the eigenvalues of $\Sq$ and $\Sk$ that fall inside the
$q_{\mathrm{eff}}$-corrected Marchenko--Pastur bulk by their mean, keeping
the spikes --- is a genuine correction for the score-aware estimator:
$+0.021$ recall at $T{=}256$ ($0.721 \to 0.742$, so 256 clipped tokens
match 512 raw ones), $+0.004$ at $T{=}512$, neutral at $T \ge 1024$
($-0.002$ at 4K); applied to PCA it \emph{hurts} at every $T$ ($-0.025$ to
$-0.042$), because PCA's variance ranking depends on exactly the bulk
structure that clipping flattens (Fig.~\ref{fig:adv}B). SAKI therefore ships with clipping enabled below $T{=}1024$ and disabled above. (The perplexity runs of Table~\ref{tab:ppl} use the plain closed form with a 512-token calibration prefix and \emph{no} clipping.) The
realized spectra are strongly spiked, so the closed form is otherwise
estimation-robust at practical calibration sizes
--- answering the static-subspace concern of \citep{ojakv} for in-domain
calibration, with domain shift still open. Fig.~\ref{fig:rmt}C gives the
diagnostic: counting Marchenko--Pastur spikes of the per-head key
correlation matrix (eigenvalues above
$\lambda_+ = (1+\sqrt{q_{\mathrm{eff}}})^2$; median 15 of 128), the
per-head SAKI$-$PCA gain \emph{anti-correlates} with spike count
($r = -0.38$; median gain $+0.061$ on low-spike heads vs.\ $+0.014$ on
high-spike). The mechanism is exactly the paper's thesis in
random-matrix form: where the key spectrum hugs the MP bulk, key variance
offers PCA no reliable signal, and query-side weighting is the only added
information --- \textbf{SAKI's advantage concentrates precisely where
PCA's objective is least informative}. The median spike count of 15 also
suggests RMT-adaptive per-head ranks well below the uniform $r{=}32$, a
principled instance of adaptive rank selection (cf.\ \citep{starkv}).

\paragraph{Ranking refinement: SAKI-Rank.} $\mathcal{L}$ is a score-MSE
surrogate; the deployed metric is top-$k$ recall, and
Prop.~\ref{prop:margin} says only errors near the margin matter. We
therefore keep the asymmetric factorization $M = B_q B_k^\top$
($B_q, B_k \in \R^{128\times 32}$), initialize it at the closed form, and
minimize a pairwise logistic loss $\log(1 + e^{\hat s_n - \hat s_p})$ over
(true top-64, hard-negative) pairs defined by the exact RoPE'd scores of
1{,}024 training queries at positions $[512, 3584)$, with a proximal
penalty $\lambda\|B - B_0\|^2$ ($\lambda{=}10^{-3}$; 300 Adam steps, lr
$10^{-3}$; the LLM is frozen; $\approx 3$\,s per head on an Apple M3 Pro
GPU). Evaluation uses the standard held-out last-512 queries. Median
recall@64 at $r{=}32$ rises from $0.799$ to $\mathbf{0.835}$ ($+0.033$
median per head; $84.3\%$ of heads improve in the independent GPU replication ($81.6\%$ in the original run); Fig.~\ref{fig:adv}C), removing
a third of PCA's remaining error ($0.748 \to 0.835$) and matching PCA's
recall at $r \approx 65$ --- half the index memory. The rare regressions
are on saturated early layers (layer~0: $0.964 \to 0.934$), where a
per-head validation gate on held-out \emph{training} queries would retain
the closed form; without any gate the two-tier story is already
\emph{SAKI-ClosedForm} (score-MSE optimal, training-free) /
\emph{SAKI-Rank} (task-aligned, one short calibration-time fit).

\emph{Does the refinement overfit its training sequence?} The factors are fitted
on one sequence, so we apply the factors trained on the literature text
\emph{unchanged} to two other sequences, each scored on its own held-out
last-512 queries (independent RTX~3090 PyTorch run, same recipe; non-fiction
taken from Chapter~IX). Median recall@64 of PCA / closed-form SAKI /
SAKI-Rank is $0.710 / 0.759 / \mathbf{0.794}$ on non-fiction and
$0.617 / 0.672 / \mathbf{0.701}$ on code: the refinement keeps its $+0.03$
gain over the closed form ($+0.032$ and $+0.031$ per head; 82\% and 76\% of
heads; 31 and 29 of 32 layers). On non-fiction the transferred SAKI-Rank
exceeds even PCA calibrated in-domain ($0.762$); on code it does not
($0.701$ vs.\ $0.744$). One training domain, one model.

\begin{figure}[t]
\centering
\includegraphics[width=\linewidth]{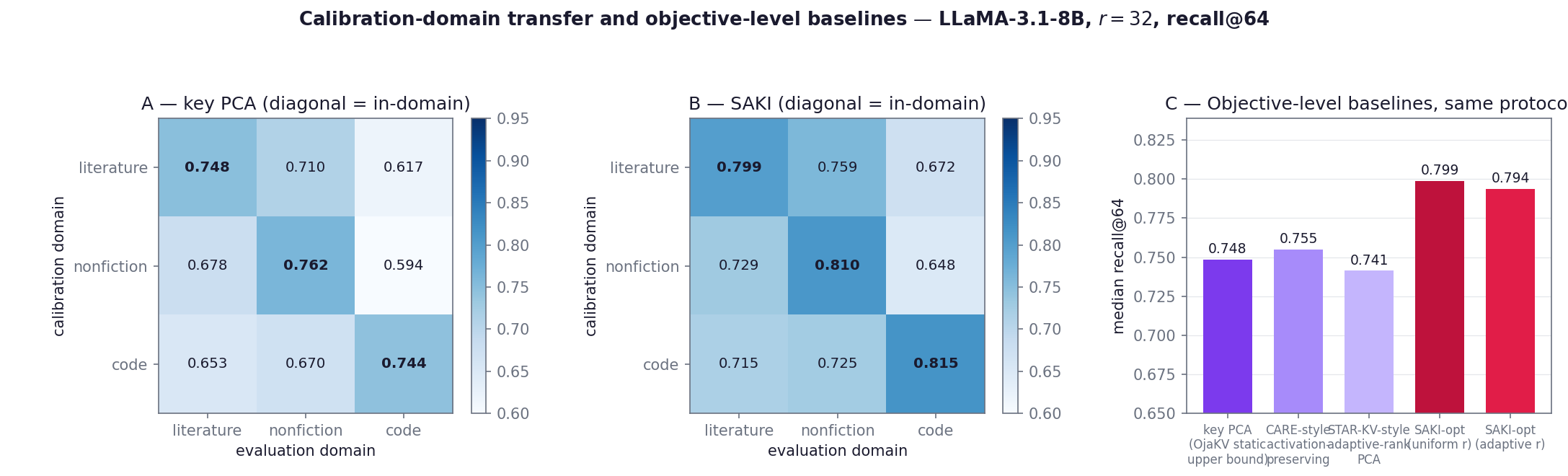}
\caption{Calibration-domain transfer and objective-level baselines,
$r{=}32$. (A)~PCA and (B)~SAKI calibrate$\to$evaluate matrices (diagonal =
in-domain). (C)~Objective-level reimplementations of the 2026 data-aware
family under the SAKI protocol.}
\label{fig:transfer}
\end{figure}

\paragraph{Calibration-domain transfer.} Static-subspace indexes are
rightly challenged on distribution shift \citep{ojakv}. We calibrate on
one domain and evaluate on another for three domains --- literature (the
default calibration text), non-fiction (Darwin's \emph{Origin of Species}),
and code (Python standard-library source; non-fiction is taken from
Chapter~IX to avoid the memorized opening) --- all nine ordered pairs, each
domain with its own exact forward pass (Table~\ref{tab:transfer};
Fig.~\ref{fig:transfer}A--B). Three facts. (i)~\emph{SAKI beats PCA in all
nine cells}, cross-domain included, by $+0.048$ to $+0.071$: the
score-aware advantage is not an artifact of matched calibration.
(ii)~\emph{Shift costs both methods about equally}: a mild shift
(literature$\leftrightarrow$non-fiction) costs SAKI $0.05$--$0.07$ recall
and a large one (prose$\leftrightarrow$code) $0.08$--$0.17$; under mild
shift, cross-domain SAKI matches or nearly matches \emph{in-domain} PCA
($0.759$ vs.\ $0.762$ on non-fiction, $0.729$ vs.\ $0.748$ on
literature), while prose$\to$code falls below it. (iii)~\emph{The
remedy follows from calibration efficiency}: since 512 tokens suffice
(Fig.~\ref{fig:rmt}A), the index can be calibrated on the prefix of the
context it will serve --- one short statistics pass, the static analogue of
OjaKV's online update. Prefix-512 in-context calibration on literature gives
$0.746$, above all but one cross-domain full-calibration SAKI value
($0.648$--$0.759$) at an eighth of the data.

\begin{table}[h]
\centering
\small
\begin{tabular}{lccc}
\toprule
calibrate $\downarrow$ / evaluate $\rightarrow$ & literature & non-fiction & code \\
\midrule
literature  & .748 / \textbf{.799} & .710 / \textbf{.759} & .617 / \textbf{.672} \\
non-fiction & .678 / \textbf{.729} & .762 / \textbf{.810} & .594 / \textbf{.648} \\
code        & .653 / \textbf{.715} & .670 / \textbf{.725} & .744 / \textbf{.815} \\
\bottomrule
\end{tabular}
\caption{Calibration-domain transfer: PCA / \textbf{SAKI}, median
recall@64, $r{=}32$, LLaMA-3.1-8B, 4K. Diagonal = in-domain. SAKI wins
every cell.}
\label{tab:transfer}
\end{table}

\paragraph{Objective-level baselines from the 2026 data-aware family.} To
compare with \citep{kvcore,care,ojakv,starkv} like with like, we reimplement their \emph{objectives} under
our protocol (their full systems add MLA conversion with healing fine-tunes,
learned thresholds with Triton kernels, quantization, and online updates ---
none of which changes the index objective). \textbf{CARE-style
activation-preserving factorization} --- truncated SVD of $W_K\Sigma_x^{1/2}$
mapped back, i.e.\ PCA of the uncentered key second moment, which is also
KV-CoRE's Frobenius-optimal data SVD --- reaches $0.755$;
\textbf{STAR-KV-style adaptive rank} (per-head ranks from a global energy
threshold tuned to a mean rank of 32; head ranks span 3--78) reaches $0.741$
for PCA and $0.794$ for SAKI; full-data static PCA ($0.748$) is the
in-domain upper bound of OjaKV's online estimate. SAKI-opt at uniform
$r{=}32$ gives $0.799$ (Fig.~\ref{fig:transfer}C). Two conclusions. First,
every reconstruction-objective variant --- centered or uncentered, uniform
or adaptive rank --- sits within $0.741$--$0.755$, and SAKI clears the best
of them by $+0.044$: the gap is to the \emph{objective class}, not to a
particular instance. Second, energy-threshold rank allocation slightly
\emph{hurts} both methods at matched budget ($-0.007$ PCA, $-0.005$ SAKI):
moving rank from the very compressible early layers (ranks as low as 3) to
deep layers trades recall where margins are wide for recall where they are
narrow, consistent with Prop.~\ref{prop:margin} --- rank should follow
score margins, not spectral energy, which is what learned allocation
\citep{starkv} and SAKI-Rank (\S\ref{sec:extensions}) target.

\paragraph{Cost.} Building the index is one pass of per-head statistics:
two $d\times d$ moments, two symmetric square roots, and one $d\times d$
SVD --- $198$\,ms per head in CPU numpy ($\approx 200$\,s for all 1{,}024
heads of an 8B model; negligible on a GPU), done once per model and
calibration set. At query time the index replaces a $d{=}128$ inner
product per cached key with an $r$-dimensional one plus a stored-mean
rank-one term; at $r{=}32$ this is cost-neutral in wall-clock on our
CPU microbenchmark (ratio $0.7$--$1.05$ across $r \in \{16,32,64\}$, these
sizes being bandwidth-bound) while the resident key codes are $r/d$ of
the full keys ($0.25\times$ at $r{=}32$). The $B_q$ query map folds into
$W_Q$ offline, so no per-query projection cost is added.

\paragraph{Memory and latency on a GPU.} Table~\ref{tab:bench} measures the
index in isolation on an RTX~3090 (synthetic data with LLaMA-3.1-8B geometry; one
decode step through 32 layers). The codes are $r/d$ of the key cache ($1.0$\,GB
against $4.0$\,GB of keys at 64K for $r{=}32$), and SAKI's equal-recall rank
($r \approx 19$--$25$, i.e.\ 59--77\% of PCA's) saves a further $0.25$--$0.4$\,GB at 64K
for an 8B model (linear in context: an extrapolated $4$--$6$\,GB per million
tokens, not measured). Per-step latency barely depends on $r$ (flat at 8K--32K,
$0.5\%$ between $r{=}24$ and $r{=}32$ at 64K; $r{=}16$ is $6\%$ faster), so the
benefit of a smaller rank is memory, not speed. Against exact attention over
the full key cache, index-based selection is slower at 8K ($7.6$ vs.\ $4.0$\,ms:
launch-bound, unfused PyTorch kernels), $1.4\times$ faster at 32K and
$2.5\times$ at 64K when everything is GPU-resident, and $1.6\times$, $5.5\times$,
$9.9\times$ faster when the KV cache lives in host memory and full attention must
stream it each step. These are attention-step microbenchmarks at batch size 1,
not end-to-end generation, and were run at the card's default 370\,W power cap
(peak draw $309$\,W, core $\le 56^\circ$C).

\begin{table}[t]
\centering
\small
\begin{tabular}{lccccccc}
\toprule
context & KV cache & codes, $r{=}24$ / $32$ & \multicolumn{2}{c}{GPU-resident step (ms)} & \multicolumn{2}{c}{host-resident step (ms)} \\
 & (GB) & (GB) & full & index ($r{=}32$) & full & index ($r{=}32$) \\
\midrule
8K & 1.00 & 0.094 / 0.125 & 3.96 & 7.55 & 46.2 & 29.4 \\
32K & 4.00 & 0.375 / 0.500 & 12.91 & 9.46 & 181.7 & 33.0 \\
64K & 8.00 & 0.750 / 1.000 & 26.23 & 10.38 & 363.2 & 36.7 \\
\bottomrule
\end{tabular}
\caption{Memory and latency of one decode step (32 layers, batch 1; RTX~3090, bf16, LLaMA-3.1-8B geometry, synthetic data, default 370\,W power cap, median of 30 repetitions, $k'{=}256$). Codes are $r/d$ of the keys; ``index'' scores the codes, selects top-$k'$ per head and runs exact attention on the selected rows; ``host-resident'' keeps the KV cache in pinned host memory, so full attention streams all of it over PCIe each step.}
\label{tab:bench}
\end{table}

\begin{figure}[t]
\centering
\includegraphics[width=\linewidth]{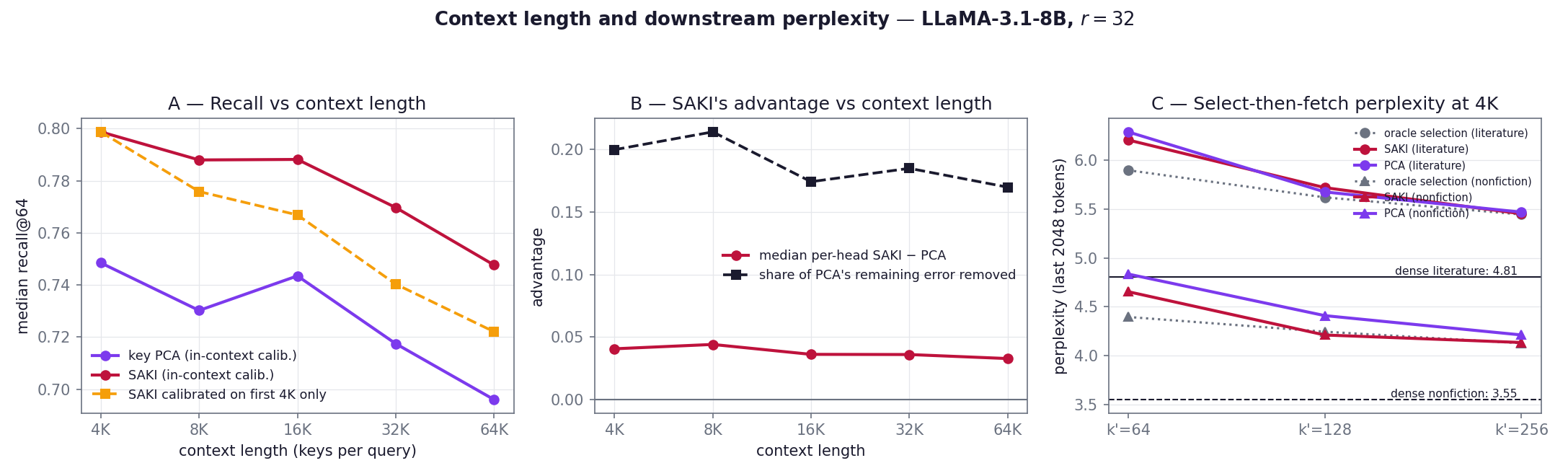}
\caption{(A)~Recall vs.\ context length from 4K to 64K ($r{=}32$; dashed:
SAKI calibrated on the first 4K only). (B)~SAKI's advantage over PCA is
flat in context length. (C)~Select-then-fetch perplexity at 4K
(Table~\ref{tab:ppl}).}
\label{fig:longctx}
\end{figure}

\paragraph{Context length.} One 65{,}536-token forward pass yields recall at
8K, 16K, 32K, and 64K (queries at each boundary, the full causal prefix as
keys), joined with the 4K result (Fig.~\ref{fig:longctx}A--B). Retrieval
gets harder for every index as the key set grows --- PCA falls from $0.748$
to $0.696$ and SAKI from $0.799$ to $0.748$ between 4K and 64K --- and
SAKI's advantage is \emph{stable} rather than compounding: median per-head
$+0.041, +0.044, +0.036, +0.036, +0.033$ at $r{=}32$ (17--21\% of PCA's
remaining error at every length; 68--77\% of heads improve), and positive at
$r{=}16$ and $r{=}64$ at every length as well. We had conjectured from the
margin condition that the gap might widen with context; the data say it
holds constant --- margins tighten for both indexes alike. Calibrating once
on the first 4K tokens and reusing the index out to 64K costs $0.01$--$0.03$
relative to in-context calibration ($0.776 \to 0.722$ vs.\ $0.788 \to
0.748$) and stays above in-context PCA at every length, so a
prefix-calibrated static index is viable across a $16\times$ context range.

\paragraph{Downstream perplexity: small, mixed, and not a claim.} Recall is
a proxy; the deployed quantity is generation quality. We run the full model
with \emph{sparse} attention in every layer and head: a rank-32 index scores
all causal keys, the top-$k'$ candidates are selected, and exact attention
(exact $q^\top k$, exact $V$, softmax over the selected set) is computed on
those candidates only --- the select-then-fetch design of \S\ref{sec:setup},
with the index calibrated in-context on the first 512 tokens.
Table~\ref{tab:ppl} gives perplexity on the last 2{,}048 of 4{,}096 tokens for
two texts; an \emph{oracle} selects by exact scores. At 4K, SAKI is better than
or equal to PCA in 5 of 6 cells and the index$-$oracle gap is at most $0.26$,
an order of magnitude smaller than the cost of sparse selection itself
(oracle$-$dense $+0.6$--$1.1$), so the selection budget $k'$ dominates
generation quality. Memorization is a confound to control: the first 4K
tokens of \emph{Origin of Species} give dense PPL $1.57$ (verbatim recall), so
non-fiction is evaluated from Chapter~IX ($3.55$). This picture does
\emph{not} extend to longer contexts or other texts. At 32K, with the
attention sink pinned (next paragraph) and a tight budget
($k' \in \{16,32,64\}$), SAKI is better than PCA in 15 of 24 (model, text,
$k'$) cells --- 11 of 12 on Qwen2.5-7B but only 4 of 12 on LLaMA-3.1-8B, where
it is worse by up to $2.6$ PPL on non-fiction (Appendix~\ref{app:ppl}) --- and
the larger recall gain of SAKI-RoPE does not produce a perplexity difference we
can resolve. The exact-score oracle is not a perplexity upper bound: per layer
it captures the most attention mass and has the lowest output error in
90--100\% of heads, yet over 32 layers it is worse than the approximate
indexes on two of the four LLaMA text windows. We therefore make no perplexity
claim.

\begin{table}[t]
\centering
\small
\begin{tabular}{llccc}
\toprule
text & dense & $k'{=}64$ & $k'{=}128$ & $k'{=}256$ \\
 & & oracle / \textbf{SAKI} / PCA & oracle / \textbf{SAKI} / PCA & oracle / \textbf{SAKI} / PCA \\
\midrule
literature  & 4.81 & 5.90 / \textbf{6.21} / 6.29 & 5.62 / \textbf{5.72} / 5.68 & 5.45 / \textbf{5.45} / 5.47 \\
non-fiction & 3.55 & 4.40 / \textbf{4.65} / 4.84 & 4.25 / \textbf{4.21} / 4.41 & 4.13 / \textbf{4.13} / 4.21 \\
\bottomrule
\end{tabular}
\caption{Perplexity of LLaMA-3.1-8B with select-then-fetch sparse attention
in all layers ($r{=}32$ index, top-$k'$ of up to 4{,}096 keys, exact
attention on the selected set), last 2{,}048 tokens. Oracle = exact-score
selection.}
\label{tab:ppl}
\end{table}

\paragraph{What is not yet shown.} No configuration reaches $0.95$ recall
at $r \le 32$ (Qwen2.5-7B comes closest at $r{=}64$: $0.914$); the recall
evidence extends to 64K context, perplexity is measured on up to four text
windows per model, and no task benchmarks are run;
LongBench/RULER is the remaining pre-submission experiment
(\S\ref{sec:limitations}).

\paragraph{Attention sinks break long-context indexes.} A failure mode that
recall hides. At 32K on LLaMA-3.1-8B (512-token calibration, $k'{=}128$),
plain SAKI is worse than PCA ($14.19$ vs.\ $10.67$; oracle $10.41$), and
calibrating on more tokens does not help --- it makes both collapse (4{,}096
calibration tokens: SAKI $64.3$, PCA $29.6$; 16{,}384 tokens: $79.7$, $35.2$).
The cause is the attention sink: position~0 is in the exact top-128 for $98\%$
of (head, query) pairs, a rank-32 index fits the bulk of the key distribution,
and a key that is an outlier of it is diluted by $1/n$ in the calibration
covariance and leaves the subspace (Table~\ref{tab:sink}: with a 512-token
calibration PCA keeps the sink 99\% of the time and SAKI 87\%; with the full
prefix 47\% and 34\%). One missing key out of 128 is invisible in recall@128 but
fatal in perplexity. Pinning the first four tokens for \emph{every} method,
inside the budget $k'$, removes the failure: at 32K SAKI and PCA reach
$10.07/10.17$ (512-token calibration) and $10.11/10.47$ (4{,}096 tokens), and
Qwen2.5-7B's PCA drops from $11.72$ to $8.81$ (SAKI $9.25 \to 8.74$). SAKI-RoPE
keeps the sink in 99\% of cases even with full-prefix calibration, without
pinning. The practical consequence is that any low-rank index evaluated at long
context must pin sink tokens; unpinned long-context numbers, including our own
preliminary 32K run, are not comparable across indexes.

\begin{table}[t]
\centering
\small
\begin{tabular}{lccc}
\toprule
calibration & PCA & SAKI & SAKI-RoPE \\
\midrule
512 tokens & 99.1\% / .689 & 86.7\% / .712 & 99.7\% / .802 \\
4{,}096 tokens & 63.0\% / .736 & 44.7\% / .760 & --- \\
full prefix & 46.7\% / .749 & 33.9\% / .791 & 99.3\% / .863 \\
\bottomrule
\end{tabular}
\caption{Attention sink at 32K (LLaMA-3.1-8B, 72 heads, $r{=}32$, $k'{=}128$): share of (head, query) pairs in which the index keeps position~0 when it is in the exact top-$k'$ / median recall@128.}
\label{tab:sink}
\end{table}

\section{Extensions}
\label{sec:extensions}
\textbf{Ranking-optimized refinement} is now measured (SAKI-Rank,
\S\ref{sec:experiments}); a per-head validation gate on held-out training
queries would remove its rare regressions on saturated early layers, and
cross-sequence transfer of the refined factors is the next test. \textbf{Per-head rank allocation} at a fixed total rank (water-filling on
the closed form's singular values, mean rank 20) gained only $+0.008$ median
recall over uniform allocation ($0.761$ vs.\ $0.753$; $0.000$ per head at the
median), so we do not pursue it.
\textbf{Query-conditioned mixtures of subspaces} (router over precomputed
$P_1 \dots P_M$) are promising but introduce training, routing, and
baseline burdens that would dilute this paper; we defer them.
\textbf{Coupling-preservation regularization}: the Schur coupling geometry
(Appendix~\ref{app:atlas}) specifies what any index must preserve and can
regularize SAKI-Rank.

\section{Limitations}
\label{sec:limitations}
Perplexity is the weakest part of the evidence, and we claim nothing from it.
With the attention sink pinned, SAKI and PCA differ by a few tenths of a
perplexity point at most in the favourable cells and by up to $2.6$ against
SAKI on LLaMA-3.1-8B non-fiction (Appendix~\ref{app:ppl}); SAKI is better in 15
of 24 tight-budget cells (11 of 12 on Qwen2.5-7B, 4 of 12 on LLaMA-3.1-8B), and
we have not explained the difference between the two models. The exact-score
oracle is not a perplexity upper bound, and recall gains --- including the larger
ones of SAKI-RoPE --- do not translate into perplexity differences we can
resolve, on a single 2{,}048-token window per text. Unpinned long-context
perplexity is unreliable for any low-rank index (\S\ref{sec:experiments}); our
preliminary unpinned 32K result (SAKI $14.19$ vs.\ PCA $10.67$ at $k'{=}128$)
should be read in that light. The gains are moderate: 3--7 recall points over
PCA for the closed form and 7--11 for SAKI-RoPE, with no task-level benchmark,
so the evidence supports a smaller index at equal recall rather than better
generation. Recall is measured from 4K to 64K context (SAKI-RoPE at 4K on all
heads of four models and at 32K on 72 and 224 heads of two); task benchmarks
(LongBench, RULER) and contexts beyond 64K are not measured. Cross-model replication is done --- four models,
uniform wins (Table~\ref{tab:xmodelsaki}) --- but the spread (Qwen 30\%
vs.\ Mistral 13\% of remaining error removed at $r{=}32$) is not yet
explained; relating it to per-model operator geometry is an open analysis.
Calibration-\emph{length} sensitivity (Fig.~\ref{fig:rmt}: 512 tokens
suffice in-domain) and calibration-\emph{domain} transfer
(Table~\ref{tab:transfer}) are both measured: shift costs recall, SAKI's
advantage over PCA survives it, and prefix calibration is the static
remedy; OjaKV-style online updates of $\Sq, \Sk$ \citep{ojakv} remain the
natural extension for contexts whose domain drifts within a single
session, which we do not test. The closed form models RoPE only through the rotation-averaged covariance (\S\ref{sec:rope}), validated at recall level and not in perplexity; the
independence model itself is validated at $r = 0.997$
(Remark~\ref{rem:indep}). Our comparison with the 2026 data-aware family
\citep{kvcore,care,ojakv,starkv} is at the objective level under a
common protocol (Fig.~\ref{fig:transfer}C); it does not reproduce their
full systems (MLA conversion and healing fine-tune in CARE; learned
thresholds, kernels, and quantization in STAR-KV; online updates in
OjaKV), whose end-to-end numbers are not comparable to a recall@64 index
metric. None of them incorporates query statistics through the score;
that remains SAKI's distinguishing design. SAKI-Rank is fitted on training queries from the same sequence as its in-domain evaluation (disjoint positions); transfer is measured from one training domain to two others, on one model, and the RMT clipping threshold ($T < 1024$) is set on one model.

\section{Conclusion}
Low-rank KV indexes have optimized weights or key variance; attention uses
neither. Deriving the expected score distortion yields a two-sided
covariance-weighted objective whose exact rank-$r$ optimum is an asymmetric,
training-free, closed-form score map, and the theory holds up: it predicts
per-head score-MSE reduction at $r = 0.997$. Making it RoPE-aware (SAKI-RoPE)
matches PCA's recall with 28--39\% of PCA's rank across four models (55--79\% without it) --- a consistent recall gain, not a step change in generation
quality: perplexity differences are small and mixed, long-context evaluation of
any low-rank index must pin the attention sink, and task-level benchmarks
remain to be run.

\appendix

\section{Full derivation of Proposition~\ref{prop:sap}}
\label{app:proof}
\paragraph{Conventions.} Column vectors $q, k \in \R^d$; $P: \R^d \to
\R^d$ linear, $\operatorname{rank}(P) \le r$; scores $s = q^\top k$,
compressed $\hat s = q^\top P k$ (centering handled below).
\paragraph{Reduction.} With $\Sq \succ 0$, $\Sk \succ 0$, substitute
$\tilde P = \Sq^{1/2} P \Sk^{1/2}$ (a bijection on rank-$\le r$ matrices,
since congruence by invertible matrices preserves rank):
$\mathcal{L}(P) = \|\Sq^{1/2}\Sk^{1/2} - \tilde P\|_F^2 = \|C - \tilde
P\|_F^2$. Eckart--Young \citep{eckart} gives $\tilde P^\star = C_r$, hence
Eq.~(\ref{eq:mr}) and $\mathcal{L}(M_r) = \sum_{i>r}\lambda_i(C)^2$.
\paragraph{Singular covariances.} If $\Sq$ or $\Sk$ is rank-deficient,
replace inverse square roots by pseudoinverse square roots; the bijection
holds on the subspace $\operatorname{range}(\Sq) \times
\operatorname{range}(\Sk)$, outside of which $\mathcal{L}$ is unaffected
by $P$, so any completion of $M_r$ by zero is optimal. In finite samples we
regularize eigenvalues below $10^{-6}\lambda_{\max}$ (all experiments);
sensitivity to this floor was not observed.
\paragraph{Centering.} Writing $k = \mu + k_c$ with $\E[k_c] = 0$ and
compressing only $k_c$: $s = q^\top k_c + q^\top \mu$; the second term is
exact at decode (store $\mu$, one inner product), so
Eq.~(\ref{eq:headspace}) applies to centered key moments and uncentered
query moments, as implemented. Compressing $k$ uncentered is the special
case $\mu = 0$.
\paragraph{Factorization identity.} With $B_q = \Sq^{-1/2} U_r
\Lambda_r^{1/2}$ and $B_k = \Sk^{-1/2} V_r \Lambda_r^{1/2}$:
$B_q B_k^\top \cdot$ transposed appropriately gives $(B_q^\top q)^\top
(B_k^\top k_c) = q^\top \Sq^{-1/2} U_r \Lambda_r V_r^\top \Sk^{-1/2} k_c =
q^\top M_r k_c$, i.e., the stored $r$-dim codes reproduce the optimal
bilinear score exactly.
\paragraph{Non-idempotence.} $M_r^2 = \Sq^{-1/2} C_r \Sk^{-1/2}
\Sq^{-1/2} C_r \Sk^{-1/2} \ne M_r$ in general (equality would require
$\Sk^{-1/2}\Sq^{-1/2}$ to act as identity on the factor range); $M_r$ is
therefore not a projector of any kind, and we do not call it one.

\section{Operator-geometry diagnosis: details}
\label{app:atlas}

\begin{figure}[h]
\centering
\includegraphics[width=\linewidth]{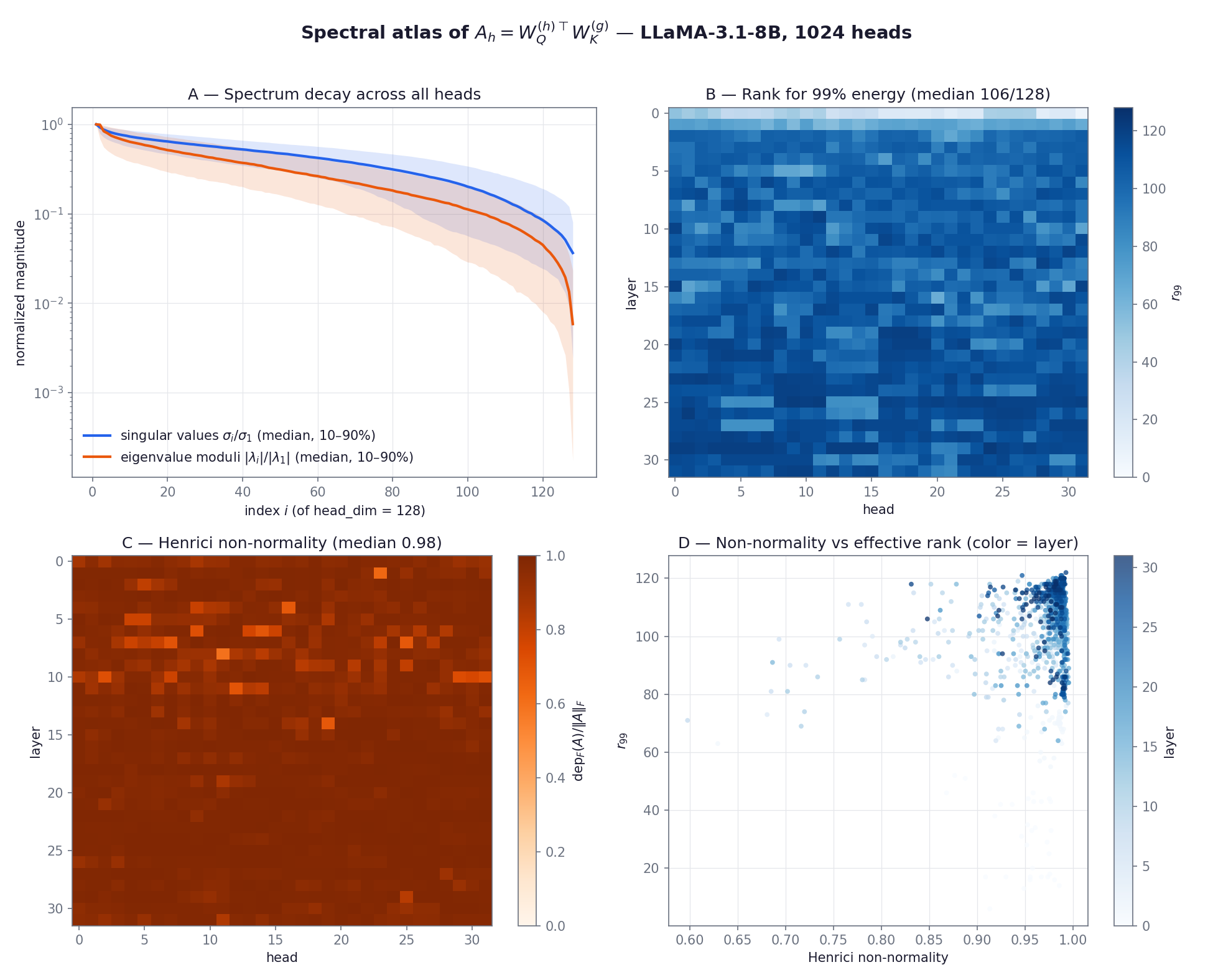}
\caption{Spectral atlas of $A_h$, all 1{,}024 heads of LLaMA-3.1-8B:
singular vs.\ eigenvalue decay (the gap is non-normality), 99\%-energy
rank, Henrici index, and their relation by layer.}
\label{fig:atlas}
\end{figure}

\begin{figure}[h]
\centering
\includegraphics[width=\linewidth]{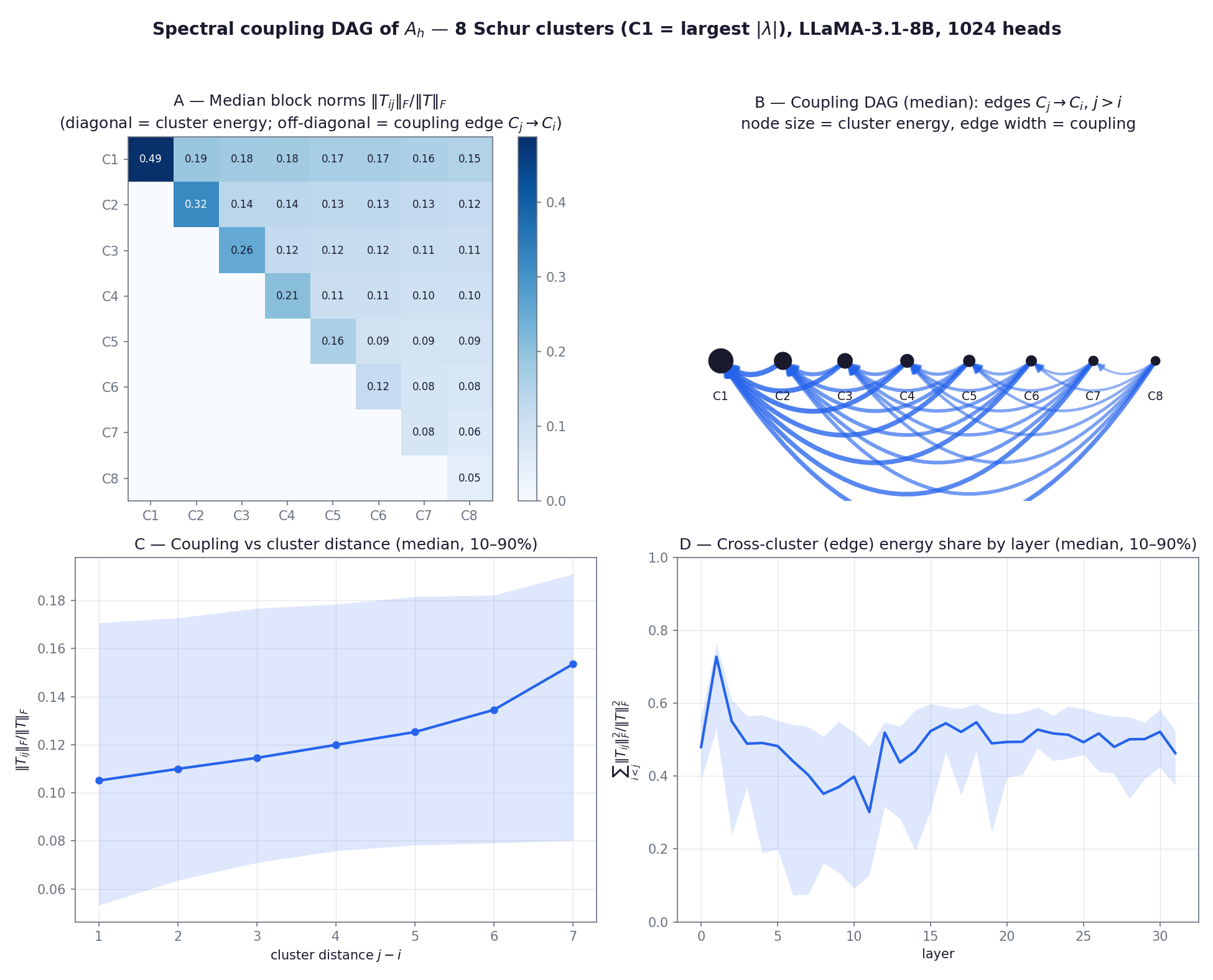}
\caption{Hub-shaped coupling geometry: modulus-ordered Schur clusters of
$A_h$ form a DAG whose couplings carry half the operator energy.}
\label{fig:dag}
\end{figure}

\begin{figure}[h]
\centering
\includegraphics[width=\linewidth]{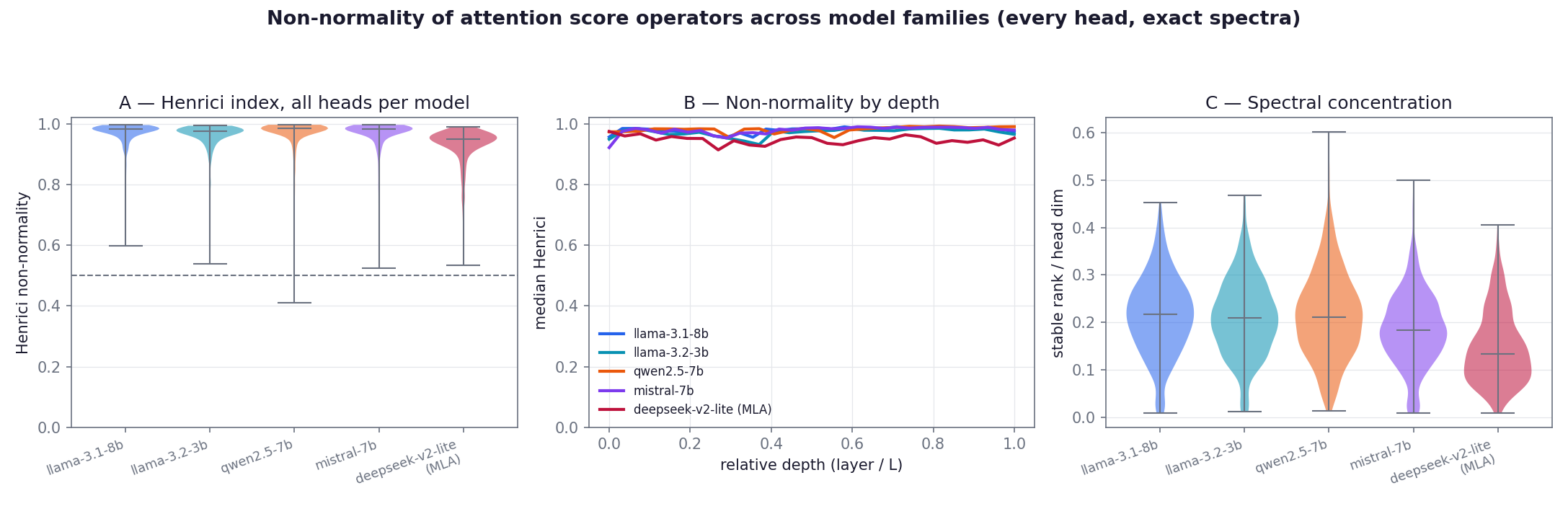}
\caption{Cross-model universality of score-operator non-normality
(every head, exact spectra, five models).}
\label{fig:xmodel}
\end{figure}

\begin{table}[h]
\centering
\begin{tabular}{lcccc}
\toprule
Model & Henrici med.\ (p10--p90) & $>0.5$ & stable rank$/d$ & complex $\lambda$ \\
\midrule
LLaMA-3.1-8B & 0.982 (0.923--0.991) & 100\% & 27.7/128 & 87\% \\
Llama-3.2-3B & 0.975 (0.916--0.986) & 100\% & 26.8/128 & 87\% \\
Qwen2.5-7B & 0.984 (0.922--0.992) & 99.9\% & 27.0/128 & 88\% \\
Mistral-7B-v0.1 & 0.982 (0.920--0.991) & 100\% & 23.5/128 & 88\% \\
DeepSeek-V2-Lite (MLA) & 0.948 (0.851--0.971) & 100\% & 25.4/192 & 88\% \\
\bottomrule
\end{tabular}
\caption{Non-normality across model families.}
\label{tab:xmodel}
\end{table}

All spectra are exact $d\times d$ computations via thin-QR reduction
($A = Q_1 M Q_2^\top$, restriction $B = MG$); the Schur coupling analysis
reorders the real Schur form of $B$ into modulus-ranked clusters
(reconstruction residual $10^{-14}$). Full statistics in the project's
\texttt{atlas\_stats.md}.

\section{Predicted vs.\ actual score-MSE reduction}
\label{app:ablation}

At $r{=}32$, per head: predicted relative score-MSE reduction
$1 - \sum_{i>32}\lambda_i(C)^2/\sum_i \lambda_i(C)^2$ vs.\ the measured
reduction on real causal (query, key) pairs (centered keys, pre-RoPE).
Across 1{,}024 heads: Pearson $r = 0.9969$; median predicted $0.969$,
median actual $0.969$; median absolute gap $0.0009$. The independence
approximation of Eq.~(\ref{eq:objective}) is near-exact for score-MSE on
natural text.

\section{Perplexity with the attention sink pinned: details}
\label{app:ppl}
Setup: 32K tokens, perplexity on the last 2{,}048, dense prefill and sparse
attention (select-then-fetch, $r{=}32$) for those queries in every layer and
head, bf16, in-context calibration on the first 512 tokens, and the first four
tokens always selected --- for the oracle, PCA and SAKI alike, and counted
inside $k'$. Table~\ref{tab:pplpin} gives SAKI$-$PCA at tight budgets on four
text windows per model (literature, non-fiction from Chapter~IX of
\emph{Origin of Species}, source code, and a second literature window). SAKI is
better in 4 of 12 LLaMA-3.1-8B cells and 11 of 12 Qwen2.5-7B cells; in the
primary window, whose results were available first, it is better in all six,
which the other windows do not reproduce.

Larger budgets on the primary window (SAKI$-$PCA at $k'{=}128/256$): LLaMA-3.1-8B
$-0.06/-0.05$ at 4K, $+0.27/+0.20$ at 16K, $-0.10/+0.06$ at 32K; Qwen2.5-7B
$-0.07/-0.03$ at 32K. All indexes are within $0.4$ PPL of the oracle at these
budgets. The RoPE-aware covariance gives $10.14/9.01$ on LLaMA-3.1-8B and
$8.73/8.60$ on Qwen2.5-7B at 32K ($k'{=}128/256$), within $0.13$ of plain SAKI
(plain: $10.07/9.14$ and $8.74/8.61$); at $k' \in \{16,32,64\}$ it is not better
than plain SAKI in most cells (LLaMA-3.1-8B, 32K: $13.74/12.25/11.24$ vs.\
$13.73/12.24/11.11$). Without pinning, fp32 reproduces the bf16 result
($14.20$ vs.\ $14.19$ for SAKI at $k'{=}128$), so precision is not the cause of the
32K reversal; RMT clipping and the RoPE-aware covariance both reduce it
($9.45$ and $10.08$) but pinning is the controlled fix.

\begin{table}[t]
\centering
\small
\begin{tabular}{lccc}
\toprule
model, text & $k'{=}16$ & $k'{=}32$ & $k'{=}64$ \\
\midrule
LLaMA, literature (primary) & $-0.02$ & $-0.16$ & $-0.22$ \\
LLaMA, non-fiction & $+2.56$ & $+1.75$ & $+1.02$ \\
LLaMA, code & $+0.33$ & $+0.25$ & $+0.10$ \\
LLaMA, literature (second) & $+0.94$ & $+0.24$ & $-0.16$ \\
Qwen, literature (primary) & $-0.42$ & $-0.25$ & $-0.16$ \\
Qwen, non-fiction & $-0.03$ & $-0.04$ & $-0.05$ \\
Qwen, code & $-0.13$ & $-0.04$ & $-0.01$ \\
Qwen, literature (second) & $+0.24$ & $-0.02$ & $-0.07$ \\
\bottomrule
\end{tabular}
\caption{SAKI$-$PCA perplexity at 32K with the first four tokens pinned for every method (negative: SAKI better). One 2{,}048-token evaluation window per row.}
\label{tab:pplpin}
\end{table}

\section{GPU experiments: setup and run details}
\label{app:gpu}
\paragraph{Hardware and software.} The long-context, Qwen2.5-7B, attention-sink
and benchmark runs use \emph{one} NVIDIA GeForce RTX~3090 (24\,GB, open kernel
module driver 595.84) of a two-GPU Ubuntu~24.04 workstation (Python~3.12, PyTorch~2.14.1 built for CUDA~13.0, cuDNN~9.24); the second card is
hidden from the process and every script refuses to start with more than one
visible device. The all-head SAKI-RoPE and cross-model comparisons at 4K, and the
32K diagnostics on 72 heads (sink retention, oracle mass, distance-resolved score
error, rank allocation), run in PyTorch on the CPU of an Apple-silicon laptop (M3~Pro).

\paragraph{Execution model.} The model is streamed layer by layer: one
transformer layer's weights (about $0.45$\,GB for the 8B models) are on the GPU at
a time, the residual stream of the $N$ tokens is kept on the device, and the
attention input of each layer is written to disk (fp16). No KV cache of the whole
model is ever built, which is why 32K--64K recall and perplexity fit on one 24\,GB
card. Captures use fp32 at 4K and bf16 at 32K and 64K; perplexities use fp32 at 4K
and bf16 from 8K up, with one fp32 control at 32K (it reproduces the bf16 result,
Appendix~\ref{app:ppl}).

\paragraph{Protocol constants.} $r{=}32$ unless stated; the index is fitted
in-context on the first $T{=}512$ tokens of the sequence (full prefix where
stated); perplexity is computed on the last 2{,}048 tokens, with dense prefill and
sparse attention only for those queries (\emph{scope eval}) from 8K up, and sparse
attention for every query (\emph{scope all}) in the 4K tables;
the selected set is the top-$k'$ keys by index score, and attention on it uses exact
$q^\top k$ and exact $V$. LLaMA-family sequences start with the BOS token; Qwen2.5
is run without one and with its q/k/v biases. Texts are public-domain
natural-language literature (primary window: character offset $10{,}000$; second
window: offset $1{,}500{,}000$), non-fiction (\emph{Origin of Species}, from
Chapter~IX, offset $518{,}313$) and source code (offset $0$); every number comes
from a single 2{,}048-token evaluation window per (model, text, length).

\paragraph{Validation.} The port is checked against the numpy/MLX reference before
any new result is trusted: token ids identical, median recall of layers 0, 16, 31
within tolerance ($0.012$; observed $\le 0.001$), dense perplexity $4.809$, and
SAKI at $k'{=}64$ $6.209$ (the final code state gives $4.809$ and $6.223$ on the CPU
path, tolerance $0.05$; the GPU path passed the same check before the long runs).

\paragraph{Run inventory.} Table~\ref{tab:gpuruns} lists what was run where. On the GPU a
perplexity run takes $5$--$25$\,s at 4K--16K and $16$--$61$\,s at 32K in bf16
($\approx 90$\,s in fp32), and a 32K capture of Qwen2.5-7B under one minute; the
all-head 4K comparison of four models takes about $70$\,s per model on the laptop CPU.

\begin{table}[t]
\centering
\small
\begin{tabular}{p{4.6cm}p{2.6cm}p{1.5cm}p{1.2cm}p{4.0cm}}
\toprule
experiment & models & context & dtype & scripts \\
\midrule
port validation & LLaMA-3.1-8B & 4K & fp32 & \texttt{validate\_port} \\
4K recall (3 domains), SAKI-Rank and transfer & LLaMA-3.1-8B & 4K & fp32 & \texttt{capture}, \texttt{saki\_rank\_torch} \\
recall vs.\ context & LLaMA-3.1-8B & 8K--64K & bf16 & \texttt{recall\_long} \\
perplexity, Table~\ref{tab:ppl} and 32K & LLaMA-3.1-8B & 4K, 32K & fp32, bf16 & \texttt{ppl\_sparse} \\
32K reversal: calibration size, precision, clipping, RoPE-aware & LLaMA-3.1-8B & 4K--32K & bf16 (fp32 control) & \texttt{diag\_ppl.sh} \\
Qwen2.5-7B: recall and perplexity & Qwen2.5-7B & 4K, 32K & bf16 & \texttt{diag\_qwen.sh}, \texttt{recall\_rot\_long} \\
pinned-sink perplexity (main, tight budgets, 3 more windows) & LLaMA-3.1-8B, Qwen2.5-7B & 4K--32K & bf16 (fp32 at 4K) & \texttt{diag\_pin.sh}, \texttt{diag\_pin\_small.sh}, \texttt{diag\_pin\_rep.sh} \\
all-head SAKI-RoPE, cross-model & four models & 4K & fp32 & \texttt{rot\_full}, \texttt{xmodel\_rot} (CPU) \\
sink retention, oracle mass, distance-resolved error, rank allocation & LLaMA-3.1-8B & 32K (72 heads), 4K & fp32 & \texttt{diag\_sink}, \texttt{diag\_oracle}, \texttt{diag\_local}, \texttt{alloc\_test} (CPU) \\
memory and latency & LLaMA geometry, synthetic & 8K--64K & bf16 & \texttt{bench\_index} \\
\bottomrule
\end{tabular}
\caption{Inventory of the GPU-side and long-context runs.}
\label{tab:gpuruns}
\end{table}

\paragraph{Thermal and power conditions.} A thermal guard pauses a job (never aborts it)
while the GPU core is above a limit and resumes below a lower one: $80/72^\circ$C for
the first LLaMA 32K runs, where several pauses occurred, and $90/85^\circ$C for the later
Qwen2.5-7B and pinned-sink runs. We did not log the board power limit for the
recall and perplexity runs and therefore report none; the latency benchmark logged the
default $370$\,W cap (peak draw $309$\,W, core $\le 56^\circ$C, no clock throttling).

\paragraph{Run integrity.} The workstation suffered two unplanned power interruptions during
the Qwen2.5-7B queue. The interrupted 32K capture left a truncated activation file; it
was detected when read, discarded, and the capture recomputed from scratch. All reported
numbers come from complete runs, and the unpinned and pinned results use the same code
path except for the pinning option.

\section{Reproducibility}
\paragraph{Independent replication.} A PyTorch/CUDA port (layer-streaming, one
RTX~3090) reproduces the Apple-silicon results: all 28 median-recall cells of the
4K--64K sweep agree to within $0.001$ (64K activations captured in bf16 there,
fp32 here), the 20 four-thousand-token perplexities to within $0.013$, and
SAKI-Rank at 4K to the third decimal ($0.835$).

{\sloppy
\paragraph{Long-context and RoPE-aware diagnostics.} The SAKI-RoPE,
cross-model, attention-sink and pinned-perplexity experiments run on the same
port (\texttt{rot\_full.py}, \texttt{xmodel\_rot.py}, \texttt{recall\_rot\_long.py},
\texttt{diag\_sink.py}, \texttt{diag\_oracle.py}, \texttt{ppl\_sparse.py}
\texttt{--pin}); Qwen2.5-7B is run with its q/k/v biases, and the first four
tokens are pinned in the selection for the oracle and both indexes alike.
}

The original pipeline runs on a single Apple-silicon workstation (float32/float64 numpy,
MLX, PyTorch-CPU); the long-context, Qwen2.5-7B, attention-sink and benchmark runs use one
RTX~3090 (Appendix~\ref{app:gpu}). Weights are range-read from public checkpoints;
calibration text is public-domain natural text.
Code: \texttt{fetch\_qk\_weights.py}, \texttt{spectral\_atlas.py},
\texttt{coupling\_dag.py}, \texttt{cross\_model\_atlas.py},
\texttt{forward\_pass.py}, \texttt{recall\_experiment.py},
\texttt{sap\_experiment.py}, \texttt{mse\_ablation.py}.
\end{document}